\documentclass[sn-mathphys]{sn-jnl}

\usepackage{bm}
\usepackage{amsfonts,amssymb}
\usepackage{makecell}
\usepackage{multirow}
\usepackage{tabularx}
\usepackage{supertabular}
\usepackage{caption}
\usepackage{bbding}
\usepackage{soul}
\usepackage{bbm}
\usepackage{hyperref}
\usepackage{bookmark}

\usepackage{color}
\usepackage{xcolor}
\newcommand{\change}[1]{\textcolor{black}{#1}}

\begin{document}

\title[Surgical-Dino: Adapter Learning of Foundation Models]{Surgical-DINO: Adapter Learning of Foundation \change{Models} for Depth Estimation in Endoscopic Surgery}

\author[1]{\fnm{Beilei} \sur{Cui}}\email{beileicui@link.cuhk.edu.hk}
\equalcont{These authors contributed equally to this work.}
\author[2]{\fnm{Mobarakol} \sur{Islam}}\email{mobarakol.islam@ucl.ac.uk}
\equalcont{These authors contributed equally to this work.}

\author[1]{\fnm{Long} \sur{Bai}}\email{b.long@link.cuhk.edu.hk}

\author*[1,3]{\fnm{Hongliang} \sur{Ren}}\email{hlren@ee.cuhk.edu.hk}

\affil[1]{\orgname{The Chinese University of Hong Kong}, \orgaddress{\city{Hong Kong}, \country{China}}}

\affil[2]{\orgdiv{Wellcome/EPSRC Centre for Interventional and Surgical Sciences (WEISS)}, \orgname{University College London}, \orgaddress{\city{London},  \country{UK}}}

\affil[3]{\orgdiv{Dept. of BME}, \orgname{National University of Singapore}, \country{Singapore}}



\abstract{\textbf{Purpose:} 
Depth estimation in robotic surgery is vital in 3D reconstruction, surgical navigation and augmented reality visualization. Although the foundation model exhibits outstanding performance in many vision tasks, including depth estimation (e.g., DINOv2), recent works observed its limitations in medical and surgical domain-specific applications. This work presents a low-ranked adaptation (LoRA) of the foundation model for surgical depth estimation.

\textbf{Methods:} 
We design a foundation model-based depth estimation method, referred to as Surgical-DINO, \change{a low-rank adaptation of the DINOv2 for depth estimation in endoscopic surgery.} We build LoRA layers and integrate them into DINO to adapt with surgery-specific domain knowledge instead of conventional fine-tuning. During training, we freeze the DINO image encoder, which shows excellent visual representation capacity, and only optimize the LoRA layers and depth decoder to integrate features from the surgical scene. 
 
\textbf{Results:} 
Our model is extensively validated on a MICCAI challenge dataset of SCARED, which is collected from da Vinci Xi endoscope surgery. We empirically show that Surgical-DINO significantly outperforms all the state-of-the-art models in endoscopic depth estimation tasks. The analysis with ablation studies has shown evidence of the remarkable effect of our LoRA layers and adaptation.

\textbf{Conclusion:} 
Surgical-DINO shed some light on the successful adaptation of the foundation models into the surgical domain for depth estimation. There is clear evidence in the results that zero-shot prediction on pre-trained weights in computer vision datasets or naive fine-tuning is not sufficient to use the foundation model in the surgical domain directly.
}

\keywords{Surgical scene understanding, Foundation models, Depth estimation, Adapter Learning}



\maketitle

\section{INTRODUCTION}
\label{sec:1}

\change{3D scene reconstruction in endoscopic surgery has a significant impact on the development of automated surgery and promotes the advancement of various downstream applications such as surgical navigation, depth perception, augmented reality, etc~\cite{zha2023endosurf, liu2019dense, wei2022unsupervised}.} However, there are still many unresolved challenges in dense depth estimation tasks within endoscopic scenes. The variability of soft tissues and occlusion by surgical tools in the surgical environment poses high demands on the model's ability to reconstruct dynamic depth maps~\cite{wang2022neural}. Recent methods have focused on utilizing binocular information to obtain disparity maps and reconstruct depth information~\cite{wang2022neural,zha2023endosurf}. However, apart from the da Vinci surgical robot system, most endoscopic surgical robot systems only consist of a monocular camera, which is a more cost-effective and easily implementable hardware solution. Therefore, precise depth estimation tasks based on monocular endoscopy are still an area that requires further exploration.

Recently, foundation models have become one of the most popular terms in the field of deep learning~\cite{kirillov2023segment,oquab2023dinov2}. Thanks to their large number of model parameters, foundation models have the ability to build long-term memory of massive training data, achieving state-of-the-art performance on various downstream tasks involving vision, text, and multimodal inputs. However, when encountering domain-specific scenarios such as surgical scenes, the predictive ability of foundation models tends to decline significantly~\cite{wang2023sam}.
Due to the limited availability of annotated data in medical scenes and insufficient computational resources, training a medical-specific foundation model from scratch poses various challenges. Therefore, there has been extensive discussion on adapting existing foundation models to different sub-domains, maximizing the utilization of existing model parameters, and fine-tuning foundation models for target application scenarios based on limited computational resources~\cite{wang2023sam,chen2023sam,wu2023self}. Chen \textit{et al.}~\cite{chen2023sam} constructed their adapter using two MLP layers and an activation function without inputting any prompt for fine-tuning the Segment Anything (SAM) model. On the other hand, Wu \textit{et al.}~\cite{wu2023self} used a simple pixel classifier as a self-prompt to achieve zero-shot segmentation based on SAM. However, the adapter layer shall slow down inference speed, and prompts cannot be directly optimized through training. Therefore, we have designed our adaptation solution based on Low-Rank Adaptation (LoRA)~\cite{hu2021lora}. LoRA adds a bypass next to the original foundation model, which performs a dimensionality reduction and then an elevation operation to simulate the intrinsic rank. When deployed in a production environment, LoRA can be introduced without introducing inference delays, and only the pre-trained model parameters need to be combined with the LoRA parameters. Therefore, LoRA can serve as an efficient adaption tool in real-world applications of foundation models.

Additionally, current works on fine-tuning vision foundation models to the medical domain have focused on common tasks such as segmentation and detection, with limited exploration in pixel-wise regression tasks like depth estimation. In this case, supervised training paradigms for visual foundation models are typically applied to common semantic understanding tasks and may not be suitable for our needs. Therefore, we have chosen DINOv2~\cite{oquab2023dinov2} as the starting point for our study in this paper. DINOv2 is a self-supervised trained foundation model for multiple vision tasks. The self-supervised training paradigm enables DINOv2 to effectively learn unified visual features, thereby requiring only customized decoders to adapt DINOv2 to various downstream visual tasks including depth estimation. Therefore, we aim to explore the fine-tuning of the DINOv2 encoder to fully utilize the pre-trained extensive parameters and benefit downstream depth estimation tasks in the surgical domain. Specifically, our key contributions and findings are:
\begin{itemize}
    \item We firstly extend the foundation model in computer vision, DINOv2, to explore its capability on medical image depth estimation problems.
    \item We present an adaptation and fine-tuning strategy of DINOv2 based on the Low-Rank Adaptation technique with low additional training costs towards the surgical image domain.
    \item \change{Our method, Surgical-DINO, is validated on two publicly available datasets and obtained superior performance over other state-of-the-art depth estimation methods for surgical images. We also investigate that the zero-shot foundation model is not yet ready for use in surgical applications, and LoRA adaptation is crucial, which outperformed naive fine-tuning.} 
\end{itemize}

\section{METHODOLOGY}
\label{sec:method}

\subsection{Preliminaries}

\subsubsection{DINOv2}
Learning pre-trained representations without regard to specific tasks has been proven extremely effective in Natural Language Processing (NLP)~\cite{raffel2020exploring}. One can use features from these pre-trained representations without fine-tuning for downstream tasks and obtain significantly better performances than those task-specific models. Oquab et al.~\cite{oquab2023dinov2} developed a similar "foundation" model, named DINOv2, in computer vision where vision features at both image level and pixel level generated from it can work without any task limitation. They proposed an automatic pipeline to build a large, curated, and dedicated image dataset and an unsupervised learning method to learn robust vision features. A ViT model~\cite{dosovitskiy2020image} with 1B parameters was trained in a discriminative self-supervised training manner and distilled into a series of smaller models that were evaluated to have surpassing ability against the best available all-purpose features on most of the benchmarks at image and pixel levels. Depth estimation task was also tested as a classical dense prediction task in computer vision by training a simple depth decoder head following DINOv2 and gained excellent performance in the general computer vision realm. The huge domain gap between medical and natural images may impede the utilization of such a foundation model thus we first attempt to develop a simple but effective adaptation method to exploit DINOv2 for the surgical domain.

\subsubsection{LoRA}

Low-Rank Adaptation (LoRA) was first proposed in~\cite{hu2021lora} to fine-tune large-scale foundation models in NLP to downstream tasks. It was inspired by the low “intrinsic dimension” of the pre-trained large model that random projection to a smaller subspace does not affect its ability to learn effectively. By injecting trainable rank decomposition matrices into each layer of the Transformer architecture and freezing the pre-trained model weights, LoRA significantly reduces the amount of trainable parameters for downstream tasks. To be specific, for a pre-trained weight matrix $W_{0}\in \mathbb{R}^{d \times k}$, LoRA utilize a low-rank decomposition to restrict its update by $ W_{0} + \Delta W = W_{0} + BA$ where $B\in \mathbb{R}^{d \times r}, A\in \mathbb{R}^{r \times k}$ with the rank $r\ll min(d,k)$. $ W_{0}$ does not receive gradient updates during the training process while only $A$ and $B$ contain trainable parameters. The modified forward pass is then described as:

\begin{equation}
h = W_{0}x + \Delta Wx = W_{0}x + BAx.
\end{equation}

This implementation can significantly reduce the memory and storage usage for training thus very suitable for fine-tuning large-scale foundational models to downstream tasks.

\subsection{Surgical-DINO}

\begin{figure}[!h]
    \centering
    \includegraphics[width=0.96\linewidth]{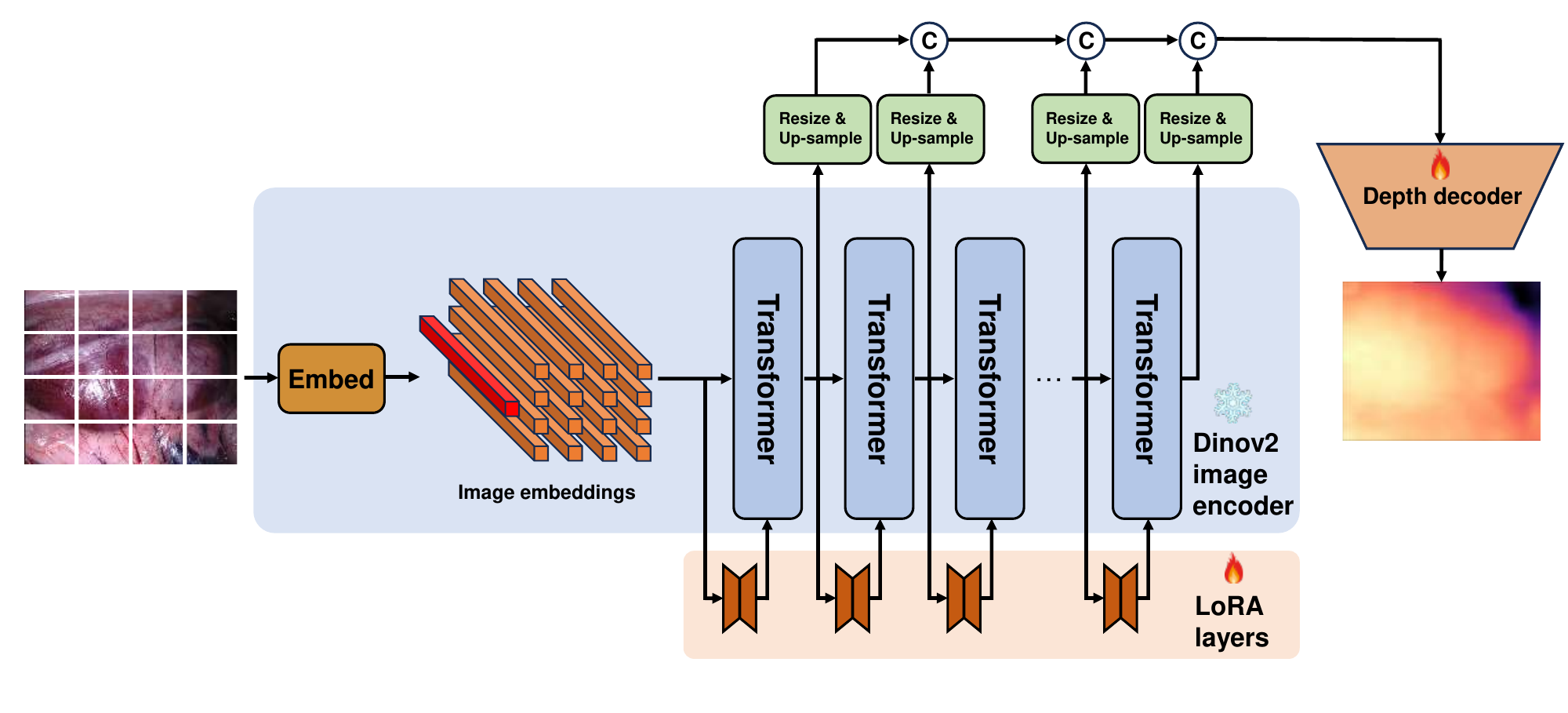}
    \caption{The proposed Surgical-DINO framework. The input image is transformed into tokens by extracting scaled-down patches
followed by a linear projection. A positional embedding and a patch-independent class token (red) are used to augment the embedding subsequently. We freeze the image encoder and add trainable LoRA layers to fine-tune the model. We extract tokens from different layers, then up-sample and concatenate them to form the embedding features. Another trainable decode head is used on top of the frozen model to estimate the final depth. }
    \label{fig: overview}
\end{figure}

As illustrated in Fig~\ref{fig: overview}, The architecture of our proposed Surgical-DINO depth estimation framework inherits from DINOv2. Given a surgical image $x\in \mathbb{R}^{H\times W\times C}$ with spatial resolution $H\times W $ and channels $C$, we aim to predict its depth map $\hat{D}\in H\times W$ as close to ground truth depth as possible. DINOv2 serves as an image encoder where images are first split into patches of size $p^2$ and then flattened with linear projection. A positional embedding is augmented for the tokens and another learnable class token is added which aggregates the global image information for subsequent missions. The image embeddings will then go through a series of Transformer blocks to generate new token representations. All parameters in the DINOv2 image encoder are frozen during training and we added additional LoRA layers to each Transformer block to capture the learnable information. These side LoRA layers, as illustrated in the previous section, compress the Transformer vision features to the low rank space and then re-project back to match the output features' channels in the frozen transformer blocks. LoRA layers in each Transformer block work independently and do not share weights. Several intermediate and the final output token representations will be resized and bi-linearly upsampled by a factor of 4 first, then concatenated to output the overall feature representation. A simple trainable Depth decoder head is utilized at the end to predict the depth map.

\subsubsection{LoRA Layers}
\begin{figure*}[]
    \centering
    \includegraphics[width=0.75\linewidth]{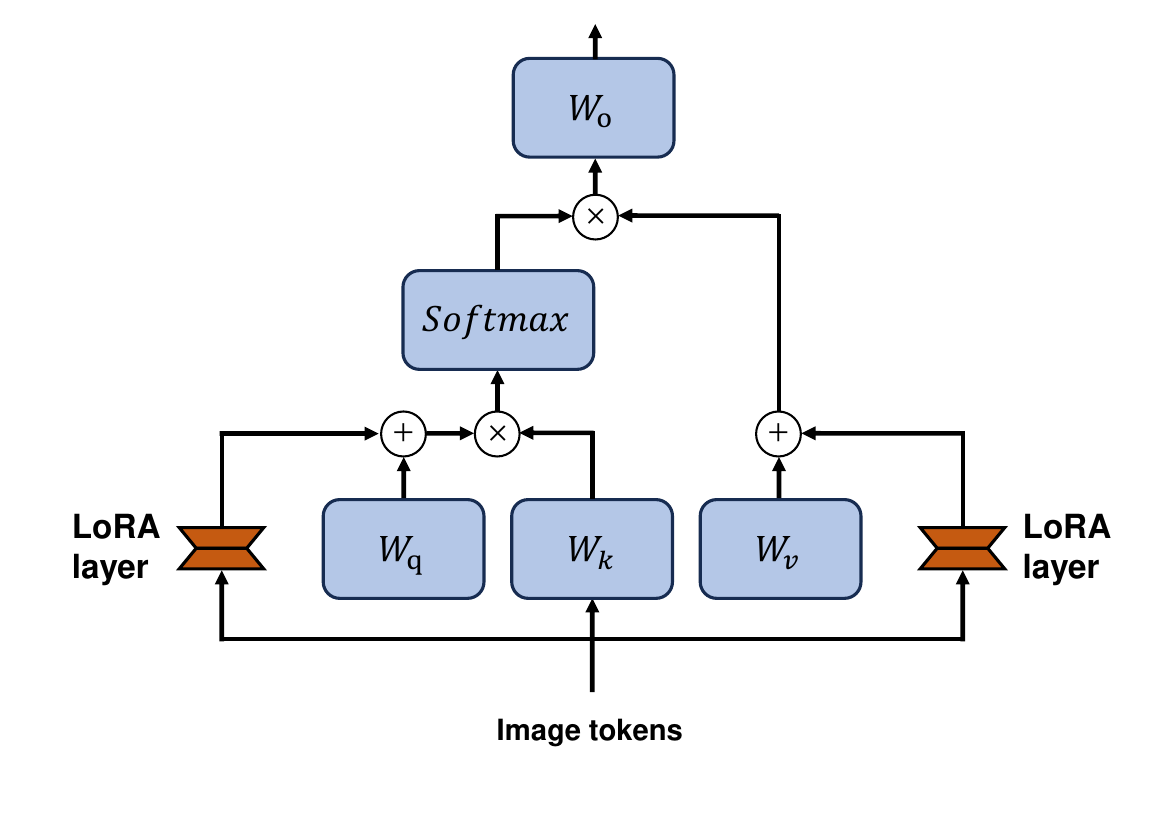}
    \caption{The LoRA design in Surgical-DINO. We apply LoRA only to $q$ and $v$ projection layers in each transformer block. $W_{q}, W_{k}, W_{v}$ and $ W_{o}$ denotes the projection layer of $q, k, v$ and $o$ respectively.}
    \label{fig: lora}
\end{figure*}

Different from fine-tuning the whole model, freezing the model and adding trainable LoRA layers will largely reduce the required memory and computation resources for training and also benefit conveniently deploying the model. The LoRA design in Surgical-DINO is presented in Fig~\ref{fig: lora}. We followed~\cite{zhang2023customized} where the low-rank approximation is only applied for $q$ and $v$ projection layers to avoid excessive influence on attention scores. With the aforementioned fundamental formulation of LoRA, for an encoded token embedding $x$, the processing of $q,k$ and $v$ projection layers within a multi-head self-attention block will become:

\begin{equation}
\begin{aligned}
Q &=\hat{W}_q a=W_q a+B_q A_q a, \\
K &=W_k a, \\
V &=\hat{W}_v a=W_v a+B_v A_v a, \\
\end{aligned}
\end{equation}
where $W_q, W_k$ and $W_v$ are frozen projection layers for $q,k$ and $v$; $A_q, B_q, A_v$ and $B_v$ are trainable LoRA layers. The self-attention mechanism remains unchanged that described by:

\begin{equation}
\operatorname{Att}(Q, K, V)=\operatorname{Softmax}\left(\frac{Q K^T}{\sqrt{C_{\text {out }}}}+B\right) V
\end{equation}
where $C_{\text {out }}$ denotes the numbers of output tokens.

\subsubsection{Network Architecture}

\textbf{Image Encoder.} The image is first separated into non-overlapping patches and then projected to image embeddings with the Embedding process. The image embeddings are a set of $t^0=\left\{t_0^0, \ldots, t_{N_p}^0\right\}, t_n^0 \in \mathbb{R}^D$ tokens, where $p$ is the patch size, $N_p=\frac{HW}{p^{2}}$, $t_0$ is the class token and $D$ represents the feature dimensions of each token. $L$ Transformers are then used to transform the image tokens into feature representations $t^l$ where $l$ denotes the output of $l$-th Transformer block. We utilized the pre-trained ViT-Base model from DINOv2 as our image encoder with 12 Transformer blocks and a feature dimension of 784.

\noindent \textbf{Depth Decoder.} We first extract the layers from $l={\left\{3,6,9,12\right\}}$, unflatten them to fit the patch resolution and up-sample tokens by a factor of 4 to increase the resolution. We treat depth prediction as a classification problem by dividing the depth range into 256 uniformly distributed bins with a linear layer to predict the depth. The predicted map is scaled to align the input resolution eventually.

\subsubsection{Loss functions}
Surgical-DINO utilizes Scale-invariant depth loss~\cite{bhat2021adabins} and Gradient loss~\cite{li2018megadepth} as the supervision constraints for the fine-tuning process. They can be described by:

\begin{equation}
\begin{aligned}
\mathcal{L}_{\text {pixel }}&=\lambda_{1}\sqrt{\frac{1}{n} \sum_i (g_i)^2-\frac{\lambda_{2}}{n^2}\left(\sum_i g_i\right)^2} \\
\mathcal{L}_{\text {grad }}&=\lambda_{3}\frac{1}{n} \sum_k \sum_i\left(  \vert\nabla_x g_i^k\vert + \vert\nabla_y g_i^k \vert\right) \\
\end{aligned}
\end{equation}
where $n$ denotes the number of valid pixels, $ g_{i}^{k} = log \tilde{d}_{i}^{k} - log d_{i}^{k}$ is the value of the log-depth difference map at position $i$ and scale $k$. $\mathcal{L}_{\text {pixel }}$ guides the network to predict truth depth while $\mathcal{L}_{\text {grad }}$ encourage the network to predict smoother gradient changes. The final loss is then described as:

\begin{equation}
\mathcal{L} = \mathcal{L}_{\text {pixel }} + \mathcal{L}_{\text {grad }}.
\end{equation}

\section{EXPERIMENT}
\label{sec:3}

\subsection{Dataset}
\label{sec:3.1}

\change{\textbf{SCARED~\cite{allan2021stereo}}} dataset is collected with a da Vinci Xi endoscope from fresh porcine cadaver abdominal anatomy and contains 35 endoscopic videos with 22950 frames. A projector is used to obtain high-quality depth maps of the scene. Each video has ground truth depth and ego-motion while we only used depth to evaluate our method. We followed the split scheme in~\cite{shao2022self} where the SCARED dataset is split into 15351, 1705, and 551 frames for the training, validation and test sets respectively. 

\noindent \change{\textbf{Hamlyn\footnote{\url{https://hamlyn.doc.ic.ac.uk/vision/}}} is a laparoscopic and endoscopic video dataset taken from various surgical procedures with challenging in vivo scenes. We followed the selection in~\cite{recasens2021endo} with 21 videos for validation.}

\subsection{Implementation Details}
\label{sec:3.2}
The framework is implemented with PyTorch on NVIDIA RTX 3090 GPU. We adopt the \change{AdamW~\cite{loshchilov2017decoupled}} optimizer with an initial learning rate of $1 \times 10^{-5}$ and weight decay of $1 \times 10^{-4}$. The batch size is set to 8 with 50 epochs in total. We can achieve our evaluation results with the following weights set: $ \lambda_{1} = 1.0, \lambda_{2} = 0.85, \lambda_{3} = 0.5$. The images are resized to $224 \times 224$ pixels. \change{We also trained our proposed model in a Self-Supervised Learning (SSL) manner with the baseline of AF-SfMLearner~\cite{shao2022self}. We replace the encoder in AF-SfMLearner with Surgical-DINO and resize the image to $224 \times 224$ pixels to fit the patch size of DINOv2.}

\subsection{Performance metrics}
We evaluate our method with five common metrics used in depth estimation problems: Abs Rel, Sq Rel, RMSE, RMSE log and $\delta$ in which lower is better for the first four metrics and larger is better for the last one. During evaluation, we re-scale the predicted depth map by a median scaling method introduced by SfM-Leaner~\cite{zhou2017unsupervised}, which can be expressed by  

\begin{equation}
\mathbf{D}_{\text {scaled }}=\left(\mathbf{D}_{\text {pred }} *\left(\text { median }\left(\mathbf{D}_{\text {gt }}\right) / \text { median }\left(\mathbf{D}_{\text {pred }}\right)\right)\right).
\end{equation}

We capped the depth map at 150 mm which can cover almost all depth values.
\subsection{Results}
\label{sec:3.3}

\begin{table}[!h]
\caption{Quantitative depth comparison on the SCARED dataset of SOTA depth estimation methods. The best results are in bold. The second-best results are underlined.}
\centering
\resizebox{\textwidth}{!}{
\begin{tabular}{c|ccccc}
\noalign{\smallskip}\hline 
Method & Abs Rel $\downarrow$ & Sq Rel $\downarrow$ & RMSE $\downarrow$ & RMSE log $\downarrow$ & $\delta \uparrow$ \\ \hline 
SfMLearner~\cite{zhou2017unsupervised} & 0.079 & 0.879 & 6.896 & 0.110 & 0.947 \\ 
Fang et al.~\cite{fang2020towards} & 0.078 & 0.794 & 6.794 & 0.109 & 0.946 \\ 
Defeat-Net~\cite{spencer2020defeat} & 0.077 & 0.792 & 6.688 & 0.108 & 0.941 \\ 
SC-SfMLearner~\cite{bian2019unsupervised} & 0.068 & 0.645 & 5.988 & 0.097 & 0.957 \\ 
Monodepth2~\cite{godard2019digging} & 0.071 & 0.590 & 5.606 & 0.094 & 0.953 \\ 
Endo-SfM~\cite{ozyoruk2021endoslam} & 0.062 & 0.606 & 5.726 & 0.093 & 0.957 \\
AF-SfMLearner~\cite{shao2022self} & \underline{0.059} & 0.435 & 4.925 & 0.082 & \underline{0.974} \\ 
DINOv2~\cite{oquab2023dinov2} (zero-shot) & 0.088 & 0.963 & 7.447 & 0.120 & 0.933 \\ 
DINOv2~\cite{oquab2023dinov2} (fine-tuned) & 0.060 & 0.459 & \underline{4.692} & \underline{0.081} & 0.963 \\
\change{Surgical-DINO SSL (Ours)} & \change{\underline{0.059}} & \change{\underline{0.427}} & \change{4.904} & \change{\underline{0.081}} & \change{0.974} \\ 
Surgical-DINO (Ours) & \textbf{0.053} & \textbf{0.377} & \textbf{4.296} & \textbf{0.074} & \textbf{0.975} \\ \hline 
\end{tabular}} 
\label{tab:main}
\end{table}

\noindent \change{\textbf{Quantitative results on SCARED.} We compared our proposed method with several SOTA self-supervised methods~\cite{zhou2017unsupervised,fang2020towards,spencer2020defeat,bian2019unsupervised,godard2019digging,ozyoruk2021endoslam,shao2022self} as well as zero-shot, self-supervised and supervised method and the results are shown in Table~\ref{tab:main}. All of these baseline methods were reproduced with the original implementation under the same dataset splits mentioned above.} The zero-shot performance of pre-trained DINOv2 is evaluated on model size ViT-Base with a same depth decoder head fine-tuned on NYU Depth V2~\cite{silberman2012indoor}. Our method obtained superior performances in all the evaluation metrics compared to all of the methods. It is worth noting that the zero-shot performance of DINOv2 has the worst results indicating that vision features and depth decoder that are highly effective in natural images are unsuitable for medical images due to the large domain gap. While the fine-tuned DINOv2 exceeds other SOTA self-supervised methods in RMSE and RMSE log, it did not gain better performance in the other three metrics proving its prediction to have more large depth errors. Only fine-tuning a depth decoder head is not enough to transfer the vision features to geometric relations within medical images. With the adaptation method of LoRA, the network is able to learn medical domain-specific vision features and relate them with depth information, thus resulting in an improvement in the estimation accuracy. 

\begin{table}[!h]
\caption{\change{Quantitative depth comparison on Hamlyn dataset. The best results are in bold. The second-best results are underlined.}}
\centering
\resizebox{\textwidth}{!}{
\begin{tabular}{c|ccccc}
\noalign{\smallskip}\hline 
\change{Method} & \change{Abs Rel $\downarrow$} & \change{Sq Rel $\downarrow$} & \change{RMSE $\downarrow$} & \change{RMSE log $\downarrow$} & \change{$\delta \uparrow$} \\ \hline 
\change{Endo-Depth-and-Motion~\cite{recasens2021endo}} & \change{0.185} & \change{5.424} & \change{16.100} & \change{0.225} & \change{0.732} \\ 
\change{AF-SfMLearner~\cite{shao2022self}} & \change{\underline{0.168}} & \change{\underline{4.440}} & \change{\underline{13.870}} & \change{\underline{0.204}} & \change{\underline{0.770}} \\ 
\change{Surgical-DINO (Ours)} & \change{\textbf{0.146}} & \change{\textbf{3.216}} & \change{\textbf{11.974}} & \change{\textbf{0.178}} & \change{\textbf{0.801}} \\ \hline 
\end{tabular}} 
\label{tab:hamlyn}
\end{table}

\noindent \change{\textbf{Quantitative results on Hamlyn.} We made zero-shot validation for our model trained on SCARED in Hamlyn dataset without any fine-tuning. For comparison, we zero-shot validate AF-SfMLeaner with their best model and obtain the results of Endo-Depth-and-Motion~\cite{recasens2021endo} by averaging the 21-fold cross-validation results trained on Hamlyn. As presented in Table~\ref{tab:hamlyn}, our method achieves superior performance against other methods, unveiling the good generalization ability across different cameras and surgical scenes. }

\begin{table}[!h]
\caption{\change{Comparison of encoder parameters, trainable parameters, trainable parameters' ratio and full model inference speed.}}
\centering
\resizebox{0.95\textwidth}{!}{
\begin{tabular}{c|cccc}
\noalign{\smallskip}\hline 
\change{Method} & \change{Params. (M) $\downarrow$} & \change{trainable Params. (M) $\downarrow$} & \change{trainable ratio (\%) $\downarrow$} & \change{Speed (ms) $\downarrow$} \\ \hline 
\change{AF-SfMLearner~\cite{shao2022self}} & \change{\textbf{11.68}} & \change{11.68} & \change{100.00} & \change{\textbf{9.9}} \\ 
\change{Surgical-DINO (Ours)} & \change{86.72} & \change{\textbf{0.14}} & \change{\textbf{0.17}} & \change{18.2} \\ \hline 
\end{tabular}} 
\label{tab:model_comple_speed}
\end{table}

\noindent \change{\textbf{Model complexity and speed evaluation.} The proposed model's parameters, trainable parameters, trainable parameters ratio and inference speed are evaluated on an NVIDIA RTX 3090 GPU compared to AF-SfMLeaner. Table~\ref{tab:model_comple_speed} shows that while Surgical-DINO has a larger amount of parameters, only a very small part of parameters are trainable making it faster to train and converge. The inference speed of Surgical-DINO is slower than AF-SfMLeaner, but still in an acceptable range for real-time applications.}

\begin{figure}[!h]
    \centering
    \includegraphics[width=0.90\linewidth]{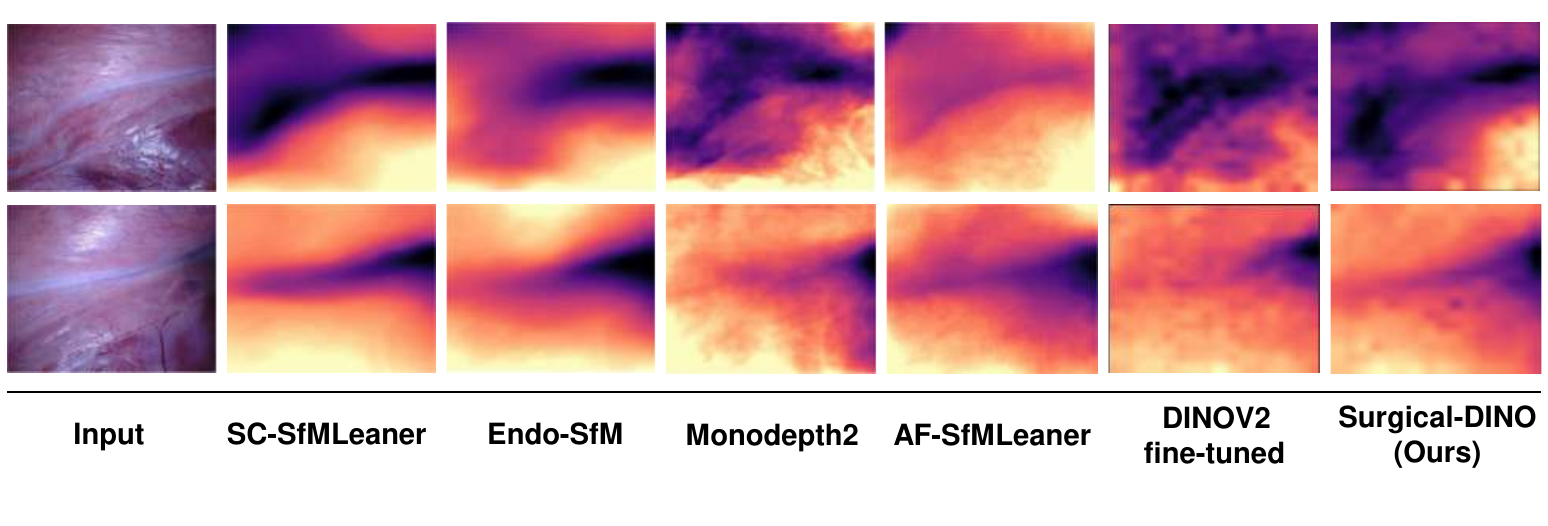}
    \caption{Qualitative depth comparison on the SCARED dataset.}
    \label{fig:qual}
\end{figure}

\noindent \textbf{Qualitative results.} We also show some qualitative results in Fig~\ref{fig:qual}. Our method can depict anatomical structure well compared to other methods. Nevertheless, the qualitative results of our proposed Surgical-DINO also has drawbacks like lack of continuity which can motivate future modification direction.

\subsection{Ablation Studies}
\label{sec:3.4}

\textbf{Effects of the rank size on the LoRA layer.} A set of comparative experiments is performed to evaluate the effects of rank size on the LoRA layer. We evaluated four different sizes of rank of the LoRA layer and the results are shown in Table~\ref{tab:abl_lorarank}. We notice that the performance of Surgical-DINO will increase with the increase of rank size within a certain low range and start to drop when the rank size exceeds a certain value. This phenomenon implies that despite being designed to utilize low-rank decomposition to make the approximation, LoRA still requires certain training parameters to fit downstream tasks. However, too many trainable parameters may mislead the original weights resulting in performance degradation.

\begin{table}[!h]
\caption{Ablation study on the rank size on the LoRA layer.}
\centering
\resizebox{0.8\textwidth}{!}{
\begin{tabular}{c|ccccc}
\noalign{\smallskip}\hline 
Rank size & Abs Rel $\downarrow$ & Sq Rel $\downarrow$ & RMSE $\downarrow$ & RMSE log $\downarrow$ & $\delta \uparrow$ \\ \hline 
1 & 0.058 & 0.389 & 4.513 & 0.081 & 0.962 \\ 
4 & \textbf{0.053} & \textbf{0.375} & \textbf{4.296} & 0.074 & \textbf{0.975} \\ 
8 & 0.053 & 0.376 & 4.324 & 0.074 & 0.974 \\ 
16 & 0.053 & 0.377 & 4.325 & \textbf{0.073} & 0.974 \\ \hline 
\end{tabular}} 
\label{tab:abl_lorarank}
\end{table}

\noindent \textbf{Effects of the size of pre-trained foundation model.} DINOV2 published four pre-trained ViT foundation models and named them by their size. Table~\ref{tab:abl_modelsize} presents the ablation study to investigate the effect of the size of the pre-trained foundation model. We discover that the performance increases with the increase of the pre-trained model size. Larger models inherently have better integration and generalization ability of vision features thus better fitting downstream tasks. But larger models are also accompanied by larger memory occupancy and training costs so we chose ViT-Base for our depth estimation method in consideration of the compromise between performance and cost.

\begin{table}[!h]
\caption{Ablation study on the size of pre-trained foundation model.}
\centering
\resizebox{0.8\textwidth}{!}{
\begin{tabular}{c|ccccc}
\noalign{\smallskip}\hline 
Model size & Abs Rel $\downarrow$ & Sq Rel $\downarrow$ & RMSE $\downarrow$ & RMSE log $\downarrow$ & $\delta \uparrow$ \\ \hline 
ViT-Small & 0.055 & 0.416 & 4.513 & 0.075 & 0.971 \\ 
ViT-Base & 0.053 & 0.377 & 4.296 & 0.074 & 0.975 \\ 
ViT-Large & 0.051 & 0.363 & 4.256 & 0.070 & 0.979 \\ 
ViT-Giant & \textbf{0.050} & \textbf{0.342} & \textbf{4.120} & \textbf{0.069} & \textbf{0.980} \\ \hline 
\end{tabular}} 
\label{tab:abl_modelsize}
\end{table}

\section{CONCLUSIONS}
\label{sec:4}

Depth estimation is a vital task in robotic surgery and benefits many downstream tasks like surgical navigation and 3D reconstruction. Vision Foundation model that captures universal vision features has been proven to be both effective and convenient in many vision tasks but yet needs more exploration in the surgical domain. We have presented Surgical-DINO, an adapter learning method that utilizes DINOv2, a vision foundation model, for surgical scene depth estimation. We design LoRA layers to fine-tune the network with a small number of additional parameters to adapt to the surgical domain. Experiments have been made on a publicly available dataset and demonstrate the superior performance of the proposed Surgical-DINO. We first explore the direction of deploying the vision foundation model to surgical depth estimation tasks and reveal its enormous potential. Future works could explore the foundation model in a supervised, self-supervised and unsupervised manner to investigate the robustness and reliability in the surgical domain.

\backmatter





\bmhead{Funding}
This work was supported by Hong Kong Research Grants Council (RGC) Collaborative Research Fund (C4026-21G), General Research Fund (GRF 14211420 \& 14203323),  Shenzhen-Hong Kong-Macau Technology Research Programme (Type C) STIC Grant SGDX20210823103535014 (202108233000303).

\bmhead{Code availability}
The source code is available at \url{https://github.com/BeileiCui/SurgicalDINO}.

\section{Declarations}

\bmhead{Conflict of interest} The authors declare that they have no conflict of interest.
\bmhead{Ethical approval} This article does not contain any studies with human participants or animals performed by any of the authors.
\bmhead{Informed consent} This articles does not contain patient data.

\bibliography{ref}


\end{document}